\documentclass[lettersize,journal]{IEEEtran}
\usepackage{amsmath,amsfonts}
\usepackage{array}
\usepackage[caption=false,font=normalsize,labelfont=sf,textfont=sf]{subfig}
\usepackage{textcomp}
\usepackage{stfloats}
\usepackage{url}
\usepackage{graphicx}
\usepackage{amssymb}
\usepackage{algpseudocode}
\usepackage{hyperref}       % hyperlinks
\usepackage{nicefrac}       % compact symbols for 1/2, etc.
\usepackage{enumitem}
\usepackage{microtype}      % microtypography
\usepackage{times}
\usepackage{latexsym}
\usepackage{xspace}
\usepackage{helvet}
\usepackage{tcolorbox}
\usepackage{verbatim}
\usepackage{anyfontsize}
\usepackage{wrapfig}
\usepackage{subcaption}
\usepackage{multicol}
\usepackage{multirow}
\usepackage{listings}
\usepackage{colortbl}
\definecolor{avgbg}{RGB}{233,245,234}
\newcommand{\tabgain}[1]{\hspace{0.05em}\raisebox{-0.45ex}{\textcolor{red}{\scriptsize$\uparrow$#1}}}
\usepackage{booktabs}
\usepackage{makecell}
\usepackage{tabularx}
\usepackage{bbm}

\def\BibTeX{{\rm B\kern-.05em{\sc i\kern-.025em b}\kern-.08em
    T\kern-.1667em\lower.7ex\hbox{E}\kern-.125emX}}
\usepackage{balance}

\begin{document}
\title{Learning Action Priors for Cross-embodiment Robot Manipulation}
\author{%
Dong~Jing$^{1,2}$, Tianqi~Zhang$^{2}$, Jiaqi~Liu$^{2}$, Jinman~Zhao$^{3}$, Zelong~Sun$^{1}$, \\Li~Erran~Li$^{4}$, Zhiwu~Lu$^{1\dagger}$, Mingyu~Ding$^{2\dagger}$\\[0.20em]
{ $^{1}$Renmin University of China, $^{2}$University of North Carolina at Chapel Hill, $^{3}$University of Toronto, $^{4}$Amazon}
\\ \vspace{-0.25in}
}

% \markboth{Journal of \LaTeX\ Class Files,~Vol.~18, No.~9, September~2020}%
% {How to Use the IEEEtran \LaTeX \ Templates}
\maketitle

\begin{abstract}
  Most Vision-Language-Action (VLA) models build on a Vision-Language Model (VLM) backbone by attaching an action module and optimizing the full policy jointly. 
  This design inherits strong visual and linguistic priors from the VLM, but leaves the action module to learn physical motion almost from scratch.
  As a result, the policy lacks an explicit motion prior, forcing early optimization to simultaneously discover temporal action dynamics and cross-modal alignment, 
  a challenge further amplified in cross-embodiment settings with heterogeneous action distributions.
  %  embodiment-specific dynamics
  % This missing motion prior slows convergence and becomes a major bottleneck, especially in cross-embodiment settings with heterogeneous action distributions.
  %
  In this work, we propose to pretrain the action module with motion priors before cross-modal VLA alignment.
  Specifically, we introduce a two-stage training framework that equips the action module with cross-embodiment temporal motion structure before VLA training begins.
  In Stage~1, a lightweight flow-matching-based encoder-decoder action module efficiently learns temporal motion structure solely from unconditioned action trajectories, without processing visual or language tokens.
  In Stage~2, this learned prior is transferred to VLA training through decoder reuse and early-stage latent distillation, aligning visual-language features with the action embedding space while still allowing end-to-end policy refinement.
  In addition, the trained encoder serves as a compact history compressor, summarizing state-action histories into a single temporal context token for history-aware modeling at negligible cost.
  Extensive experiments across 13 diverse cross-embodiment tasks on both simulated and real-world platforms validate the effectiveness of our approach.
  Compared with VLA training without action priors, our model achieves faster convergence, higher success rates, and substantially stronger performance on data-scarce real-world tasks.
  Moreover, scaling up the action data in Stage~1 yields a more generalizable action prior that directly improves downstream VLA performance.
  % especially in long-horizon scenarios where the history context token is critical.
  % The learned encoder also serves as a compact history compressor, injecting temporal state-action context into the VLM at negligible computational cost.
\end{abstract}
\vspace{-0.1in}
\vspace{-0.1in}

\begin{figure*}[!t]
  \centering
  \includegraphics[width=0.99\linewidth]{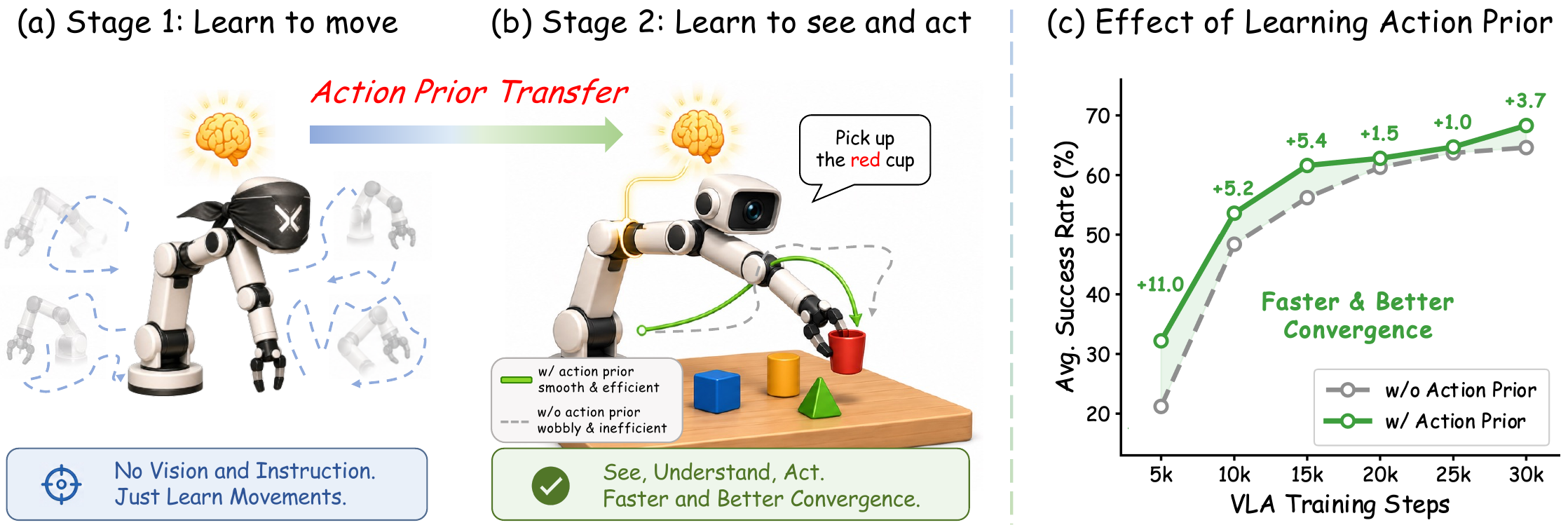}
  \caption{Illustration of motivation: a policy should first learn to move, and then learn to see and act. (a) In Stage~1, the action module is trained purely on action trajectories, without any visual observation or language instruction, to efficiently acquire a general action prior. (b) In Stage~2, the prior is transferred to VLA training, enabling model to perceive scenes and follow instructions with an action module that already knows how to move. With the action prior, the policy produces smooth trajectories. Without it, the trajectories are unstable. (c) The action prior accelerates VLA training and improves performance.}
  \label{intro_teaser}
  \vspace{-0.1in}
  \end{figure*}
\section{Introduction}
\label{sec:intro}

In recent years, VLAs have rapidly emerged as a dominant paradigm for robotic manipulation. These models build upon foundation VLMs~\cite{liu2023llava,chen2024internvl,bai2025qwen2,achiam2023gpt} or video generation models~\cite{du2023unipi,black2024susie,cheang2024gr2,agarwal2026cosmos,ye2026world}, which encode rich multi-modal semantic knowledge. Such semantic foundations allow VLAs to connect high-level language instruction and visual observations with low-level physical execution. This paradigm has achieved remarkable success across diverse robotic platforms, such as tabletop manipulators~\cite{brohan2022rt,brohan2023rt,kim24openvla}, quadruped robots~\cite{miki2022perceptive,hoeller2024anymal}, and humanoid systems~\cite{fu2024humanplus,bjorck2025gr00t,cheang2025gr}. 
These advances open a credible path toward generalist robotic policy. 
However, the success of current VLAs is still largely driven by visual and linguistic priors inherited from large-scale foundation-model training, while the action generation module itself usually receives no comparable prior over physical motion.

Missing action priors create a key imbalance in the standard VLA training pipeline. Existing methods typically append an action module to a foundation VLM and optimize the whole model jointly via imitation learning~\cite{brohan2023rt,black2024pi_0,li2024cogact}.
While the VLM already encodes rich semantic structure, the action module is often initialized from random weights or from parameters transferred from unrelated modalities. 
It therefore leaves the action module to learn physical motion and temporal dynamics from scratch, creating a bottleneck in modeling robot action distributions.
% It therefore enters training with little knowledge of the temporal structure and distribution of robot actions.
%
As a result, early VLA training must learn both the action distribution and the cross-modal alignment between perception, language, and control at the same time. The gradients produced by such an under-trained action head can be unstable and poorly structured, interfering with the foundation backbone and slowing convergence. This issue becomes more pronounced in cross-embodiment settings~\cite{o2024open,wang2024hpt}, where different robots exhibit substantially different tasks, action spaces, and motion distributions.

To address this imbalance, we introduce \textit{action prior learning before cross-modal VLA alignment}, targeting only the action module and using only action data. As shown in Figure~\ref{intro_teaser}(a), the action module first learns to move from unconditioned action trajectories, without any visual observation or language instruction. Much like a blindfolded apprentice practicing basic movements, it focuses on the intrinsic temporal structure of physical motion instead of learning action dynamics and cross-modal alignment simultaneously. The learned prior is then transferred to Stage~2, where the VLA learns to perceive the scene and follow instructions on top of an action module that already encodes plausible motion patterns, as illustrated in Figure~\ref{intro_teaser}(b). This decoupled training process yields smoother trajectories and faster convergence, as summarized in Figure~\ref{intro_teaser}(b,c), because VLA training begins from a structured action module rather than an untrained action head.

We instantiate this action prior with an encoder-decoder action module. The encoder is designed to extract the motion semantics that emerge across an action sequence, rather than from isolated atomic actions. A single action usually records only an infinitesimal state change over a short time interval, while meaningful behaviors such as translating, rotating, or grasping arise from the continuous composition of actions over a temporal horizon. The encoder therefore compresses an entire action chunk into a compact latent embedding that captures its temporal dynamics and macroscopic motion structure. Conditioned on this embedding, the decoder reconstructs the original action chunk via a flow-matching objective, which provides a flexible way to model the continuous action distribution and its uncertainty.

We transfer this learned prior to VLA training through three tightly connected uses of the learned action module: \emph{decoder reuse}, \emph{early-stage latent distillation}, and \emph{history compression}. At the action-output side, the decoder that already models plausible robot motion from Stage 1 is reused as the VLA action head. At the representation side, the encoder provides a latent target for early-stage distillation: given the ground-truth action chunk, it produces a structured embedding that supervises the VLM's predicted action feature and quickly anchors it to the learned motion latent space. This alignment is gradually relaxed, allowing later training to refine the policy end-to-end without being permanently constrained by the encoder. Finally, the encoder also serves as a compact history compressor, mapping past state-action trajectories into a single latent token that injects temporal context into the VLM. Together, these mechanisms turn the action prior from a standalone reconstruction model into a practical initialization, alignment signal, and memory interface for VLA training, leading to smoother rollouts and faster convergence as shown in Figure~\ref{intro_teaser}(b,c).

We evaluate our method on a demanding cross-embodiment setup that spans 13 tasks across two simulated benchmarks, LIBERO and RoboCasa, and a real-world Franka platform.
All models are trained jointly on heterogeneous embodiments and evaluated without environment-specific fine-tuning.
Our model has several appealing benefits.
\textbf{1)} Action prior learning provides the most direct benefits expected from the proposed design: \textbf{faster convergence and higher success rates} over from-scratch VLA training.
\textbf{2)} The gains are especially pronounced on data-scarce long-tail tasks.
In real-world tasks with a small number of demonstrations, the action prior improves \textbf{long-tail stability} by making the policy behave more stably and decisively, mitigating the uneven convergence in which data-rich simulated tasks dominate while rare real-world behaviors remain underfit.
\textbf{3)} We further observe \textbf{favorable scaling with action-only data}: increasing the amount of unconditioned action data in Stage~1 yields a broader action prior, and it transfers directly to downstream VLA training.
\textbf{4)} History integration brings huge gains for \textbf{long-horizon tasks} via enlarged action receptive field.
The learned encoder effectively extracts and compresses past state-action trajectories into a compact latent token, expanding the temporal context visible to the policy.
% This enlarged action receptive field supports more stable decision making and reduces hesitation at critical execution points.

Our main contributions are summarized as follows:

\begin{enumerate}
    \item We identify the absence of action priors as a structural bottleneck in VLA training, and introduce a two-stage framework that efficiently learns action-centric robot motion before cross-modal policy alignment.
    \item We introduce a flow-matching encoder-decoder action module that learns compact motion-aware embeddings and transfers them as action priors for VLA training. This structure further yields a history context token for efficient history-aware decision making.
    \item We validate the method across 13 cross-embodiment tasks in simulation and the real world, showing faster convergence, higher success rates, improved long-tail stability, and favorable scaling with more action data.
\end{enumerate}

\section{Related Work}

\subsection{Vision-Language-Action Models}
Vision-Language-Action models (VLAs)~\cite{yuan2026qwen,deng2025graspvla,dreamvla25,cheang2025gr,bjorck2025gr00t,univla,mu2024embodiedgpt,starvla2025,jing2026tempovla} have rapidly emerged as a dominant paradigm for robotic manipulation by leveraging the semantic priors of foundation Vision-Language Models (VLMs)~\cite{bai2025qwen3,steiner2024paligemma}.
Early efforts such as RT-2~\cite{brohan2023rt} demonstrated that directly fine-tuning a VLM on robotic data can produce effective visuomotor policies.
Subsequent works have explored diverse architectural designs.
OpenVLA~\cite{kim24openvla} and its efficient variant OpenVLA-OFT~\cite{kim2025openvla-oft} adopt an open-source VLM backbone with action tokenization for scalable training.
On the action generation side, Diffusion Policy~\cite{chi2023diffusion} and ACT~\cite{Zhao2023ACT} demonstrate that expressive generative models can effectively capture multi-modal action distributions, and RDT-1B~\cite{liu2024rdt} scales diffusion-based action generation to a foundation model for bimanual manipulation.
Building on these ideas, CogACT~\cite{li2024cogact} appends a diffusion-based action module to a VLM, generating actions conditioned on the VLM's latent features.
$\pi_0$~\cite{black2024pi_0} employs a mixture-of-transformers architecture with a flow-matching action head, and $\pi_0$-FAST~\cite{pi0-fast} further improves efficiency through discrete action tokenization.
Other lines of work focus on enriching spatial reasoning~\cite{qu2025spatialvla,li2025spatial,tang2025geomanip}, incorporating chain-of-thought reasoning~\cite{cot-vla,zawalski2024robotic,tang2025incentivizing}, visual trace prompting~\cite{zheng2024tracevla}, or introducing world model capabilities from Video Generatation Models~\cite{cen2025worldvla,dreamvla25,li2026causal,yuan2026fast,kim2026cosmos}.
Despite these advances, most VLA models still rely on action modules that lack motion priors comparable to the semantic priors inherited by the VLM backbone.
This creates a huge gap between the foundation model's visual-language knowledge and the action head's understanding of physical motion.

\subsection{Cross-Embodied Robot Manipulation}
Training a single policy across multiple robot embodiments promises broader generalization and improved data efficiency.
This vision has been fueled by the emergence of large-scale, multi-embodiment datasets.
Open X-Embodiment~\cite{o2024open} consolidated over one million trajectories from 22 robot platforms across 60+ datasets, establishing the first large-scale cross-embodiment benchmark.
BridgeData V2~\cite{walke2023bridgedata} provides over 60k trajectories of diverse manipulation behaviors for scalable, open-vocabulary policy learning.
DROID~\cite{khazatsky2024droid} collects 76k demonstrations across 564 scenes by 50 operators spanning three continents.
AgiBot World~\cite{bu2025agibot} further scales to over one million trajectories across 217 tasks collected from 100 real robots.
RoboMIND~\cite{wu2024robomind} contributes 107k trajectories across four embodiments with a unified collection protocol.
RoboTwin~\cite{chen2025robotwin2} provides a scalable simulation benchmark for bimanual manipulation with domain randomization.
LIBERO~\cite{liu2023libero} and RoboCasa~\cite{nasiriany2024robocasa} offer diverse simulation benchmarks for lifelong learning and everyday household tasks, respectively.

On the modeling side, Octo~\cite{team2024octo} adopts a transformer architecture with modality-specific tokenizers to handle heterogeneous observation and action spaces.
GR00T N1~\cite{bjorck2025gr00t} leverages a diffusion transformer for humanoid robot control across diverse embodiments.
X-VLA~\cite{zheng2025xvla} introduces soft-prompt tokens and a unified action space to disambiguate between embodiments, enabling scalable cross-embodiment training.
UniVLA~\cite{univla} learns task-centric latent actions to achieve embodiment-agnostic policy transfer.
Universal Actions~\cite{zheng2025universal} proposes a standardized action representation that maps heterogeneous action spaces into a common format.
A key challenge in cross-embodiment settings is that the diverse action distributions across platforms significantly increase the difficulty of action prediction.
This motivates mechanisms that expose the policy to heterogeneous action distributions before cross-modal policy learning, rather than leaving the action module to absorb embodiment differences only through downstream VLA supervision.

\subsection{Action Prior Learning}
Action prior learning is related to methods that introduce action-side structure before or during policy learning.
ACT~\cite{Zhao2023ACT} uses a CVAE~\cite{sohn2015learning} encoder-decoder over action sequences, but its encoder is observation-conditioned and trained end-to-end inside the full policy.
APT~\cite{xu2026apt} follows a staged VA-to-VLA recipe, first learning a vision-action expert and then transferring it to instruction-conditioned VLA training.
It is related to our sequential training setup, but it is mainly designed to improve instruction following.

Another line of work learns action-like representations from visual trajectories or videos.
LAPA~\cite{ye2024latent} discretizes latent actions between consecutive video frames with a VQ-VAE~\cite{van2017neural}.
IGOR~\cite{chen2024igor} uses image-goal representations as atomic control units, while Moto~\cite{chen2024moto} learns latent motion tokens through a video-generation-style objective.
These methods use frame-to-frame visual change as supervision, making the learned representation closely tied to visual prediction.
This design is attractive for web-scale videos~\cite{egodex,punamiya2026egoverse,grauman2022ego4d}, but the resulting tokens often describe how images evolve rather than directly modeling the robot's motor distribution.

World-model-based approaches further use visual dynamics to learn latent action or motion structure.
VLA-JEPA~\cite{ma2025vlajepa} predicts future state representations with a latent world model, and LAWM~\cite{wu2025lawm} learns latent actions through self-supervised world modeling on unlabeled videos.
AdaWorld~\cite{gao2025adaworld} learns adaptable world models with latent actions, while DreamDojo~\cite{gao2026dreamdojo} scales generalist robot world modeling to large-scale human videos.
DynaMo~\cite{cui2024dynamo} learns dynamics-aware visual features through in-domain dynamics prediction.
These methods place greater emphasis on predicting forward dynamics in the visual space.
Action-like variables are then recovered by inverting these dynamics, which differs from directly modeling the robot action distribution.
In contrast, our goal is to fit the action space itself as a reusable prior for downstream VLA training.
% Their objectives therefore emphasize predictive structure in observation space, while our goal is to fit the action space itself as a reusable prior for downstream VLA training.

Overall, these approaches show that latent action or dynamics representations can improve robot policy learning.
Most of them, however, start from visual data, future-state prediction, or vision-action policy learning.
Our setting instead uses robot action trajectories directly, without visual observations or language instructions.
This isolates the intrinsic statistics of physical motion and keeps the learning stage lightweight.

\section{Methodology}

\subsection{Preliminary and Task Formulation}
\noindent\textbf{Standard VLA Formulation.} Vision-Language-Action models (VLAs) formulate robotic manipulation as a conditional action generation task.
Given visual observations $o_t$ and a language instruction $l_t$ at timestep $t$, the objective is to predict a future action chunk $a = [a_t, a_{t+1}, \dots, a_{t+H-1}]$, where $H$ denotes the action chunk horizon~\cite{Zhao2023ACT,jing2025mixture}. 
Most VLAs are built upon foundation Vision-Language Models (VLMs).
Let $\theta$ denote the parameters of the backbone.
To introduce the action modality, an action module parameterized by $\phi$ is appended, and the full parameters $\{\theta, \phi \}$ are optimized jointly via a behavioral cloning (BC) objective over the embodiment dataset $\mathcal{D}$:

\begin{equation}
    \min_{\theta,\phi} \mathbb{E}_{(o, l, a) \sim \mathcal{D}} [ \mathcal{L}_{BC}(f_{\theta,\phi}(o, l), a) ],
\end{equation}
where $f_{\theta,\phi}$ represents the entire VLA network, and $\mathcal{L}_{BC}$ denotes a generic behavioral cloning objective for action prediction, instantiated as regression, diffusion, or flow matching depending on the action head.

\vspace{0.05in}
\noindent\textbf{Bottleneck.}
However, a critical bottleneck arises during the initial phase of training.
Because the action module parameters $\phi$ are typically initialized randomly or transferred from a disjoint modality such as vision or language, the action module enters training with little knowledge of the temporal structure and distribution of robot motion.
Early optimization must solve two coupled problems at once: the action module has to learn a continuous action distribution, while the VLM backbone has to align visual-language representations with this still-evolving action space.
When the action head is under-trained, the gradients $\nabla_\theta \mathcal{L}_{BC}$ backpropagated to the VLM can be poorly structured and unstable, forcing the foundation backbone to adapt to a moving action target and slowing convergence.

\begin{figure*}[!t]
\centering
\includegraphics[width=0.99\linewidth]{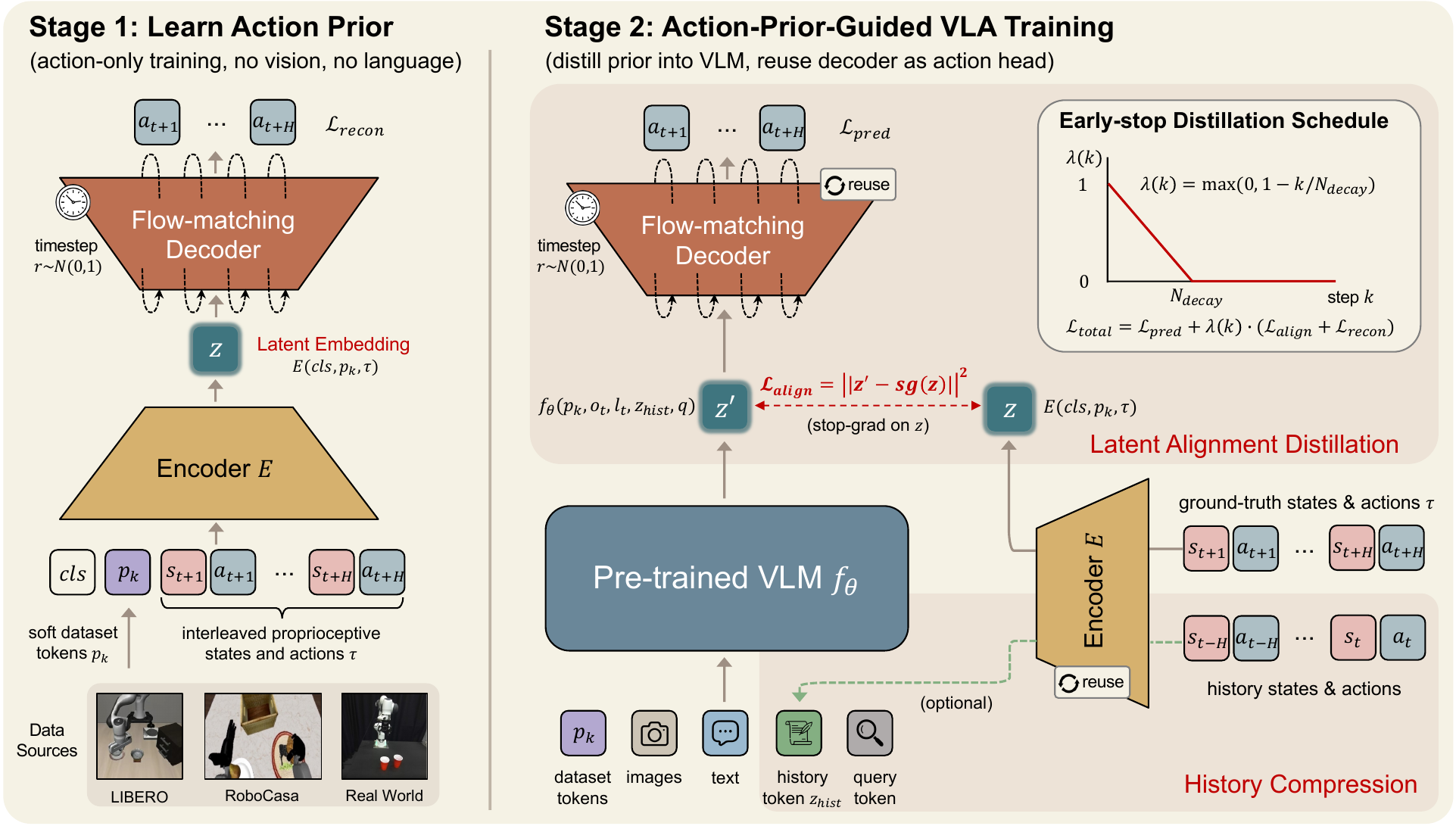}
\vspace{0.05in}
\caption{Architecture of our proposed framework. In Stage~1, interleaved state-action sequences and dataset-specific soft-prompt tokens are fed into the encoder to produce a latent action embedding $z$. The flow-matching decoder then reconstructs the original actions conditioned on $z$. 
In Stage~2, the VLM processes soft data tokens, visual observations, language instructions, and a learnable query token to predict $z'$. The predicted embedding $z'$ is aligned with the encoder's $z$ through a decaying distillation loss and passed to the Stage~1-initialized decoder for action generation. 
The same encoder compresses historical trajectories into a single token $z_{hist}$, injecting temporal context into the VLM at negligible cost.}
\vspace{-0.05in}
\label{fig:framework}
\end{figure*}

% \vspace{-0.05in}
\subsection{Learning Action Prior}

\noindent\textbf{Motivation.} 
While large-scale visual-language training successfully establishes robust visual and linguistic representations in VLMs, the action modality itself contains rich intrinsic prior knowledge that remains largely underexploited.
As shown in Figure~\ref{fig:framework}, we therefore introduce a dedicated action prior learning stage that trains the action module independently on low-level trajectory data through a reconstruction objective, enabling it to acquire an \textit{Action Prior} before it is coupled with visual-language inputs.
This stage is action-centric because no visual observations or language instructions are used.

\vspace{0.05in}
\noindent\textbf{Definition of Action Prior.}
We define the Action Prior along two fundamental dimensions:
1) \textit{Distribution Modeling}: the capability to fit the underlying continuous action distribution.
2) \textit{Structural Abstraction}: the ability to compress temporally extended action chunks into meaningful latent embeddings.
Crucially, isolated atomic actions carry little semantic meaning on their own.
It is the continuous composition of actions across a temporal horizon that gives rise to meaningful physical motions, such as translating, rotating, and grasping.
The action module should therefore capture these temporal dynamics and project them into a regularized latent space.

\vspace{0.05in}

\noindent\textbf{Formulation.}
To learn this prior, we introduce a standalone action module comprising an encoder $E_{\phi_{enc}}$ and a decoder $D_{\phi_{dec}}$, with joint parameters $\phi = \{\phi_{enc}, \phi_{dec}\}$.
Given trajectory data $\tau$ with action chunk $a$ from the embodiment dataset $\mathcal{D}$, we optimize $\phi$ by minimizing an action reconstruction objective without any visual-language conditioning:
\begin{equation}
\phi^{*} = \arg \min_{\phi} \mathbb{E}_{(\tau, a) \sim \mathcal{D}} [\mathcal{L}_{recon}(D_{\phi_{dec}}(E_{\phi_{enc}}(\tau)), a)],
\end{equation}
where $\phi^{*} = \{\phi^{*}_{enc}, \phi^{*}_{dec}\}$ denotes the parameters that successfully capture the action prior.

The learned action module is then reused in Stage~2 in three complementary ways: the decoder initializes the VLA action head, the encoder provides a structured latent target for early-stage distillation, and also compresses historical state-action trajectories into a compact context token.
% By introducing this latent alignment as a form of distillation, we seamlessly transfer the structural action prior into $\theta$, bridging the representation gap and drastically accelerating cross-modal convergence.

\subsection{Encoder-Decoder Action Modeling}

We instantiate the Stage~1 action module with a Transformer-based encoder-decoder architecture.
As illustrated in Figure~\ref{fig:framework}, this module is trained independently on low-level trajectory data via the reconstruction objective, decoupling the acquisition of the action prior from visual-language alignment.

\vspace{0.05in}
\noindent\textbf{Action Encoder.} 
The action encoder $E_{\phi_{enc}}$ compresses a low-level trajectory into a single dense latent embedding that captures the global temporal structure of motion.
To construct the input sequence, we interleave the robot's proprioceptive states $s$ and actions $a$ into a sequence $\tau = [s_t, a_t, s_{t+1}, a_{t+1}, \ldots]$.
Actions describe local motion increments, while proprioceptive states anchor these increments to the robot's body configuration.

In general cross-embodiment settings, state and action spaces differ in dimensionality, semantics, and control conventions across platforms.
To mitigate this heterogeneity, following X-VLA~\cite{zheng2025xvla}, we prepend learnable dataset embeddings $p_k$ to indicate the source embodiment and disambiguate embodiment-specific action semantics.

A learnable summary token $cls$ is also prepended, and the full encoding process is formulated as:
\begin{equation}
    z = E_{\phi_{enc}}(cls, p_k, \tau),
\end{equation}
where $z$ is the $\ell_2$-normalized output feature of $cls$, serving as the latent action embedding.

By aggregating the entire sequence into a single token, $z$ is explicitly trained to capture the global dependencies and structural dynamics of the action chunk, representing the macroscopic motion semantics that emerge from the continuous combination of actions over a temporal horizon.

\vspace{0.05in}
\noindent\textbf{Flow-Matching-based Action Decoder.} 
Conditioned on the latent embedding $z$, the decoder $D_{\phi_{dec}}$ reconstructs the original action chunk. 
We adopt a flow-matching paradigm for this purpose, which provides a flexible generative objective for continuous and potentially multi-modal action distributions.

Sharing a similar Transformer architecture as the encoder, the decoder takes the latent embedding $z$ alongside noise-corrupted actions $x_r$ as input.
We define a probability path between standard Gaussian noise $\epsilon \sim \mathcal{N}(0, I)$ and the ground-truth actions $a$, where the noise-corrupted actions at training flow-matching timestep $r \in [0, 1]$ is constructed via linear interpolation:
\begin{equation}
    x_r = ra + (1 - r)\epsilon.
\end{equation}
The target velocity field that drives noises toward clean actions is $v_r = a - \epsilon$.

Following $\pi_{0.5}$~\cite{pi05}, the timestep $r$ is integrated into the normalization layer by projecting it into modulation weights.
The decoder is optimized as a velocity prediction network using the flow-matching objective:
\begin{equation}
    \mathcal{L}_{recon} = ||D_{\phi_{dec}}(x_r, r, z) - v_r||^2
\end{equation}
By minimizing this objective over low-level trajectory data without visual-language conditioning, the decoder learns a robust and generalizable action distribution prior, laying a stable foundation for the subsequent VLA training.

\subsection{VLA Training with Action Prior Distillation}

Having acquired a robust action prior through the independent learning stage, we transfer it to VLA training by reusing the learned action module and distilling its latent structure into the VLM.

\vspace{0.05in}
\noindent\textbf{VLA Architecture.} 
As shown in Figure~\ref{fig:framework}, we adopt a foundation VLM as the backbone $f_\theta$.
At timestep $t$, the VLM takes the soft-prompt dataset embeddings $p_k$, current visual observations $o_t$, the language instruction $l_t$, and a learnable query token $q$ as inputs. 
The output feature corresponding to $q$ is extracted and $\ell_2$-normalized as the predicted action latent embedding $z'$:
\begin{equation}
    z' = f_\theta(p_k, o_t, l_t, q).
\end{equation}

Instead of learning the action head from scratch, we initialize the VLA action decoder with the Stage~1 decoder and jointly optimize it during VLA training.
Conditioned on $z'$, the decoder predicts the velocity field for action generation, yielding the primary prediction loss:
\begin{equation}
    \mathcal{L}_{pred} = \| D_{\phi_{dec}}(x_r, r, z') - v_r \|^2.
\end{equation}

\vspace{0.05in}
\noindent\textbf{Latent Alignment Distillation.}
The Stage~1 encoder provides a direct way to supervise the VLM in the learned action latent space.
Given the ground-truth trajectory, the action encoder $E_{\phi_{enc}}$ extracts a structured latent embedding $z$, which serves as a teacher target for the VLM-predicted embedding $z'$.
The alignment loss is defined as:
\begin{equation}
    \mathcal{L}_{align} = \| z' - \mathrm{sg}(z) \|^2,
\end{equation}
where $\mathrm{sg}(\cdot)$ denotes the stop-gradient operator.
This prevents the alignment objective from pulling the encoder toward the current VLM prediction, while still encouraging the VLM to enter the motion-aware latent space learned in Stage~1.
To maintain the coherence of this latent space during joint VLA optimization, we also keep the reconstruction objective $\mathcal{L}_{recon}$ on the encoder-decoder branch.
This auxiliary loss uses the same flow-matching formulation as in Stage~1 with the ground-truth embedding $z$, preserving the action latent geometry rather than directly supervising the VLM prediction.

\vspace{0.05in}
\noindent\textbf{Early-step Distillation Strategy.}
While $\mathcal{L}_{align}$ provides critical early-stage guidance, maintaining this strict constraint throughout training could overly restrict the VLM's representational flexibility and hinder end-to-end performance refinement.
We therefore propose an early-step distillation strategy, introducing a monotonically decaying weight $\lambda(k)$ at training step $k$:
\begin{equation}
    \lambda(k) = \max\left(0, 1 - \frac{k}{N_{decay}}\right),
\end{equation}
where $N_{decay}$ is a predefined decay threshold.

% Therefore, we propose an early-step distillation strategy. 
% We introduce a monotonically decaying weight $\lambda(k)$ for the distillation losses at training step $k$:
% \begin{equation}
%     \lambda(k) = max(0,1-\frac{k}{N_{decay}}),
% \end{equation}
% where $N_{decay}$ is a predefined threshold step.

The total VLA training objective is formulated as:
\begin{equation}
\label{equ:loss}
    \mathcal{L} = \mathcal{L}_{pred} + \lambda(k)(\mathcal{L}_{align} + \mathcal{L}_{recon})
\end{equation}
During early training, the distillation losses anchor the VLM to a physically meaningful action representation subspace.
As $k$ exceeds $N_{decay}$, these constraints are phased out, and the model transitions to standard end-to-end VLA optimization driven by $\mathcal{L}_{pred}$.

\subsection{Temporal Context Compression with the Action Encoder}
\label{sec:history} 

\noindent\textbf{Motivation.}
Sequential manipulation often requires temporal context beyond the current observation.
Past states and actions reveal whether the robot has approached, grasped, hesitated, or already completed part of a motion.
However, directly appending raw state-action history to the VLM input is inefficient and poorly matched to the VLM's image-language token space.
Long histories introduce substantial token overhead, and their low-level numerical values force the VLM to learn temporal abstraction from scratch.

\vspace{0.05in}
\noindent\textbf{Encoder-based History Compression.}
We address this issue by reusing the action encoder as a compact history compressor.
Because the encoder is trained to summarize trajectory structure in Stage~1, the same mechanism can also extract motion-aware context from past state-action sequences.
Let $\tau_{hist} = [s_{t-H_h}, a_{t-H_h}, \ldots, s_{t-1}, a_{t-1}]$ denote the historical trajectory over a horizon $H_h$.
As illustrated in Figure~\ref{fig:framework}, we pass $\tau_{hist}$ through the learned encoder to obtain a single historical latent token $z_{hist}$:
\begin{equation}
    z_{hist} = E_{\phi_{enc}} (cls, p_k, \tau_{hist}).
\end{equation}
This token is then inserted into the VLM input, updating the forward pass to condition on temporal context:
\begin{equation}
    z' = f_\theta(p_k, o_t, l_t, z_{hist}, q).
\end{equation}

This design injects historical information in a form that is both compact and semantically structured.
Instead of exposing the VLM to raw numerical trajectories, $z_{hist}$ carries motion dynamics already organized by the action prior.
Meanwhile, compressing the entire history into one token provides a wider temporal view with negligible computational overhead.

\vspace{0.05in}
\noindent\textbf{History-conditioned Training.}
During training, we compute the full objective in Equation~\ref{equ:loss} under both the history-conditioned and history-free settings, and average two losses:
\begin{equation}
    \mathcal{L}_{total} = (\mathcal{L}(z_{hist}) + \mathcal{L}(\emptyset)) / 2,
\end{equation}
where $\mathcal{L}(z_{hist})$ and $\mathcal{L}(\emptyset)$ denote the training objective conditioned with and without the historical embedding, respectively.
This dual-mode training encourages the policy to benefit from temporal context when available, while avoiding over-reliance on the history token.

\section{Experiment}

\subsection{Experimental Setup}

\begin{figure*}[!t]
\centering
\includegraphics[width=0.99\linewidth]{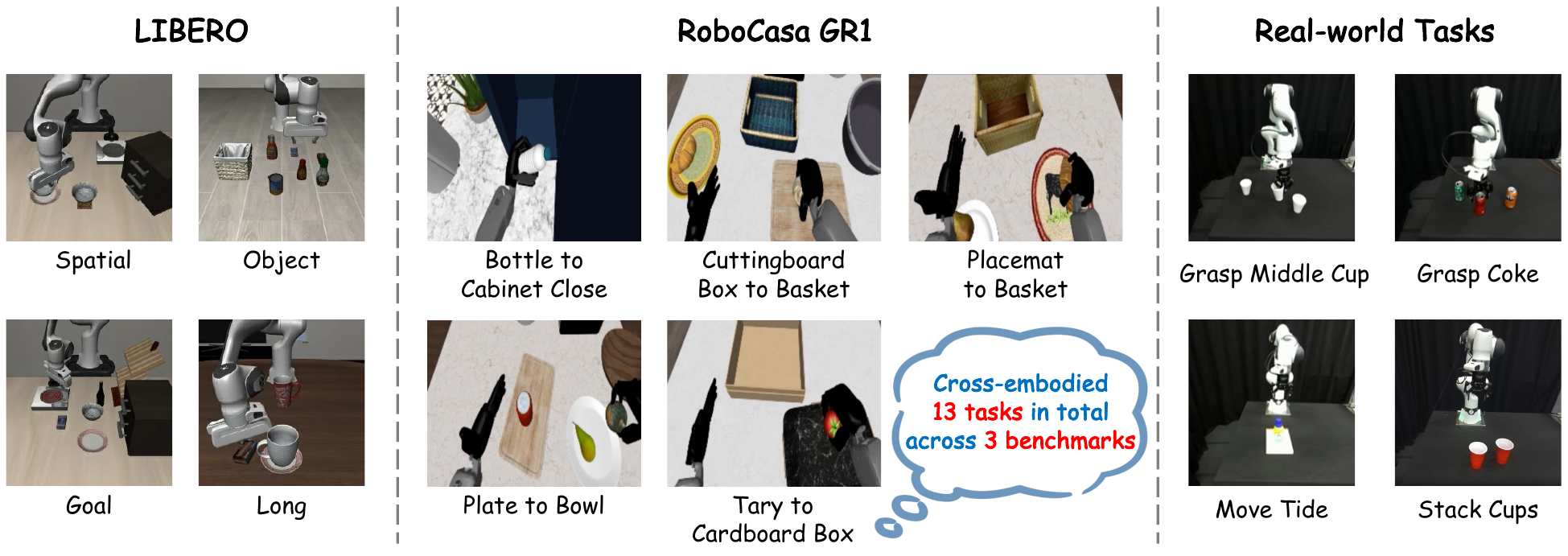}
\caption{Overview of the 13 cross-embodiment tasks across LIBERO, RoboCasa GR1, and real-world Franka manipulation. The suite covers heterogeneous embodiments, action-state spaces, scenes, and task distributions.}
\label{fig:tasks}
\end{figure*}

\vspace{0.05in}
\noindent\textbf{Cross-Embodiment Data Setting.}
As shown in Figure~\ref{fig:tasks}, our experiments cover 13 tasks across two simulated benchmarks (LIBERO and RoboCasa) and a real-world Franka platform.
The three environments differ substantially in embodiment type, action and state representation, scene composition, and task distribution, making joint policy learning considerably more demanding than single-embodiment training.
All models are trained on the full cross-embodiment mixture and evaluated on each benchmark without environment-specific fine-tuning.

\textit{LIBERO~\cite{liu2023libero}.}
LIBERO comprises four task suites: Spatial, Object, Goal, and Long.
Each suite contains 10 tasks and 500 demonstrations for training, and is designed to probe generalization across different spatial layouts, object categories, task goals, and long-horizon sequences, respectively.
The four task suites require progressively longer trajectories.
During evaluation, we test each task for 50 episodes with a fixed random seed, resulting in 500 episodes per suite. At each inference step, the model predicts an action chunk and the first 5 actions are executed by default, after which a new prediction is made. This cycle repeats until the task is completed or the maximum step limit is reached.

\textit{RoboCasa~\cite{nasiriany2024robocasa}.}
RoboCasa is a large-scale simulation framework featuring diverse manipulation environments built around the GR1 humanoid robot platform.
We select 5 tabletop tasks from the RoboCasa GR1 benchmark, each involving structured rearrangement operations across varied objects and receptacle configurations.
For each task, we use 200 demonstration trajectories, yielding 1,000 trajectories in total for training.
Each task is evaluated for 50 episodes, totaling 250 episodes across all five tasks. The first 10 actions of each predicted action chunk are executed.

\textit{Real-World Tasks.}
Beyond simulation, we design 4 pick-and-place tasks on a real Franka robot arm for evaluation.
The training demonstrations are collected via teleoperation, with 50 demonstration trajectories per task.
Each task involves a distinct set of objects and requires qualitatively different motion trajectories, ensuring diversity in both scene configuration and manipulation style.
Each task is evaluated for 10 trials, and the first 8 actions of each predicted action chunk are executed.

The full training mixture contains approximately 565k state-action transitions.
The four real-world tasks contribute only 42{,}967 frames, accounting for about 7.6\% of the mixture, which makes them a natural long-tail testbed.

\vspace{0.05in}
\noindent\textbf{Unified Action-State Space.}
To enable joint training across heterogeneous embodiments, we define a unified action space of 37 dimensions and a unified state space of 74 dimensions, as detailed in Table~\ref{tab:unified_repr}.
Each embodiment's native action and state vectors are mapped into corresponding slots of the unified vector, with all unused dimensions zero-padded.
Specifically, LIBERO uses 7-DoF end-effector delta actions (consisting of 3-DoF position, 3-DoF rotation, 1-DoF gripper) mapped to the right-arm slot, RoboCasa GR1 uses 29-DoF actions covering both arms, hands, and waist, and real-world Franka uses 8-DoF joint-space actions.
For states, LIBERO provides 8-DoF end-effector readings, real-world Franka provides 8-DoF joint angles, and GR1 provides full-body joint states expanded to 58 dimensions via sine-cosine transformation.
This slot-based design prevents interference between embodiments, while the learnable dataset tokens $p_k$ further disambiguate embodiment-specific patterns.

\begin{table}[!t]
\centering
\caption{Unified 37-D action and 74-D state representation across embodiments. Unused slots are zero-padded.}
\label{tab:unified_repr}
\renewcommand{\arraystretch}{1.12}
\resizebox{0.92\columnwidth}{!}{%
\begin{tabular}{@{}cl c ccc@{}}
\toprule
 & Component & Dim & LIBERO & GR1 & Real \\
\midrule
\multirow{6}{*}{\rotatebox[origin=c]{90}{\scriptsize\textbf{Action 37D}}}
 & Left arm & [0:7] & & \checkmark & \\
 & Right arm & [7:14] & \checkmark & \checkmark & \\
 & Left hand & [14:20] & & \checkmark & \\
 & Right hand & [20:26] & & \checkmark & \\
 & Waist & [26:29] & & \checkmark & \\
 & Joint angles + gripper & [29:37] & & & \checkmark \\
\midrule
\multirow{3}{*}{\rotatebox[origin=c]{90}{\scriptsize\textbf{State 74D}}}
 & EE pose + gripper & [0:8] & \checkmark & & \\
 & Joint angles + gripper & [8:16] & & & \checkmark \\
 & Full body  & [16:74] & & \checkmark & \\
\bottomrule
\end{tabular}%
}
\end{table}

\vspace{0.05in}
\noindent\textbf{Baselines and Variants.}
We evaluate our method against representative VLA architectures and internal variants under the same cross-embodiment training protocol.
For a controlled comparison, all models are trained within the same StarVLA codebase~\cite{community2026starvla} on the identical cross-embodiment dataset, without using externally trained VLA checkpoints.
To ensure fairness and enhance cross-embodiment capability, all methods use learnable soft-prompt dataset tokens $p_k$.

\textit{No Action Prior.}
This baseline adopts the same encoder-decoder action architecture and flow-matching decoder as our full method, but entirely removes Stage~1 action prior learning.
The action module is initialized with generic LLM parameters as shown in Table~\ref{tab:hyperparams} and trained jointly with the VLM using only the prediction loss $\mathcal{L}_{pred}$.
This configuration is structurally similar to CogACT~\cite{li2024cogact}, and therefore also serves as a direct comparison with that approach.

\textit{Action-State Prior.}
This variant introduces Stage~1 action prior learning with both action and proprioceptive state trajectories, and transfers the learned module to Stage~2 through decoder initialization and latent distillation.

\textit{Action-State Prior + History.}
This is our full model, which additionally injects the encoder-compressed history token into the VLM during Stage~2.

\textit{GR00T}~\cite{bjorck2025gr00t}.
GR00T employs a diffusion transformer~\cite{sohl2015deep,ho2020denoising,peebles2023scalable} as the action prediction head.
The VLM backbone extracts visual-language features that condition the diffusion transformer through cross-attention layers to generate actions via iterative denoising.
We re-train GR00T under the same data and optimization protocol for a controlled comparison.

\textit{$\pi_{0.5}$}~\cite{pi05}.
$\pi_{0.5}$ features a mixture-of-transformers architecture comprising a VLM for visual-language understanding and a lightweight action expert for action generation.
The two modules interact through cross-attention, and the action expert generates actions via flow-matching.
We adopt PaliGemma-2B as the VLM backbone and Gemma-300M as the action expert.

\begin{table}[!t]
  \centering
  \caption{Training configuration for Stage~1 action prior learning and Stage~2 VLA training.}
  \label{tab:hyperparams}
  \footnotesize
  \renewcommand{\arraystretch}{0.95}
  \setlength{\tabcolsep}{16pt}
  \begin{tabular}{@{}lcc@{}}
  \toprule
   & Stage 1 & Stage 2 \\
  \midrule
  \multicolumn{1}{@{}>{\columncolor{avgbg}[0pt][\tabcolsep]}l}{Backbone} &
  \multicolumn{1}{>{\columncolor{avgbg}[\tabcolsep][\tabcolsep]}c}{Qwen3-0.6B} &
  \multicolumn{1}{>{\columncolor{avgbg}[\tabcolsep][0pt]}c@{}}{Qwen3-VL-2B} \\
  Chunk size $H$ & 15 & 15 \\
  \multicolumn{1}{@{}>{\columncolor{avgbg}[0pt][\tabcolsep]}l}{Batch size} &
  \multicolumn{1}{>{\columncolor{avgbg}[\tabcolsep][\tabcolsep]}c}{2048} &
  \multicolumn{1}{>{\columncolor{avgbg}[\tabcolsep][0pt]}c@{}}{256} \\
  Optimizer & AdamW & AdamW \\
  \multicolumn{1}{@{}>{\columncolor{avgbg}[0pt][\tabcolsep]}l}{Learning rate} &
  \multicolumn{1}{>{\columncolor{avgbg}[\tabcolsep][\tabcolsep]}c}{$1\!\times\!10^{-4}$} &
  \multicolumn{1}{>{\columncolor{avgbg}[\tabcolsep][0pt]}c@{}}{$2.5\!\times\!10^{-5}$} \\
  LR scheduler & Warmup-Cosine & Warmup-Cosine \\
  Min LR & $5\!\times\!10^{-7}$ & $2.5\!\times\!10^{-8}$ \\
  \multicolumn{1}{@{}>{\columncolor{avgbg}[0pt][\tabcolsep]}l}{Training steps} &
  \multicolumn{1}{>{\columncolor{avgbg}[\tabcolsep][\tabcolsep]}c}{5{,}000} &
  \multicolumn{1}{>{\columncolor{avgbg}[\tabcolsep][0pt]}c@{}}{50{,}000} \\
  Warmup steps & 1{,}000 & 5{,}000 \\
  $N_\text{decay}$ & -- & 5{,}000 \\
  GPUs & 8$\times$H200 & 8$\times$H200 \\
  \multicolumn{1}{@{}>{\columncolor{avgbg}[0pt][\tabcolsep]}l}{Training time} &
  \multicolumn{1}{>{\columncolor{avgbg}[\tabcolsep][\tabcolsep]}c}{2 hours} &
  \multicolumn{1}{>{\columncolor{avgbg}[\tabcolsep][0pt]}c@{}}{20 hours} \\
  \bottomrule
  \end{tabular}
\end{table}
  
\vspace{0.05in}
\noindent\textbf{Training hyperparameters.}
Table~\ref{tab:hyperparams} summarizes the training configurations for both stages.
In Stage~1, the action encoder and decoder are initialized with Qwen3-0.6B~\cite{yang2024qwen2} and trained solely on unconditioned cross-embodiment state-action trajectories to acquire the action prior.
In Stage~2, we adopt Qwen3-VL-2B~\cite{Qwen3-VL} as the VLM backbone and jointly optimize the VLM and action module for 50{,}000 steps across $13$ cross-embodiement tasks.
The alignment and reconstruction losses are linearly decayed to zero over the first $N_{decay} = 5{,}000$ steps.
All experiments are conducted on 8$\times$NVIDIA H200 GPUs with BF16 mixed precision.

\subsection{Main Results on Cross-Embodiment Manipulation}

\vspace{0.05in}
\noindent\textbf{Overall Performance.}
Table~\ref{tab:main} summarizes the comparison across all 13 cross-embodiment tasks.
We focus on whether action prior learning improves VLA training under two central challenges: heterogeneous embodiments and uneven data distributions, especially the long-tail setting represented by real-world tasks.
The results show three consistent benefits: stronger overall performance, substantially better long-tail real-world behavior, and complementary gains from history compression.

\vspace{0.05in}
\noindent\textbf{Action prior learning improves both simulation and real-world performance.}
Action-State Prior improves the overall average from 55.3\% to 64.9\%, and adding history further raises it to 68.0\%.
It also improves the simulation average from 64.3\% to 66.5\%, with the history variant reaching 68.8\%.
The gain is especially pronounced on the real-world Franka tasks, where the average success rate increases from 35.0\% to 61.3\% with the action prior and to 66.3\% with history enabled.
Our full model also clearly outperforms GR00T and $\pi_{0.5}$, which reach 48.6\% and 53.8\% overall, respectively.
Notably, $\pi_{0.5}$ remains competitive in simulation but collapses on the real robot, whereas our approach maintains strong performance across both domains.

\vspace{0.05in}
\noindent\textbf{The action prior helps most on long-tail real-world tasks.}
Each real-world task contains only 50 teleoperated demonstrations; together, the real-world subset contributes only 42{,}967 frames, or 7.6\% of all training frames, forming the data-scarce tail of the mixture.
The gains from the action prior are largest on this tail: Grasp Coke improves from 5\% to 35\%, and Stack Cups improves from 25\% to 75\%.
In contrast, the competing methods struggle on these tasks, suggesting that the action prior is particularly valuable when downstream VLA supervision is sparse and unevenly represented.
$\pi_{0.5}$ provides a clear example: despite its strong simulation performance, it fails on every Stack Cups trial.
We analyze the behavioral causes in Section~\ref{sec:qualitative}.

\vspace{0.05in}
\noindent\textbf{History compression provides a complementary boost.}
Enabling history raises Grasp Coke from 35\% to 50\% and Stack Cups from 75\% to 80\%, lifting the overall average from 64.9\% to 68.0\%.
We attribute this gain to the enlarged temporal receptive field provided by the compressed history token, which helps reduce hesitation around critical decision points, as analyzed below.
This suggests that the encoder-compressed history is most useful when the policy must decide whether to keep aligning or commit to the terminal action.

\begin{table*}[!t]
\centering
\caption{Comparison with state-of-the-art methods on 13 cross-embodiment tasks.
Success rates (\%) are reported.
All methods use the same StarVLA codebase and training data.
Sim Avg. denotes the average success rate over the 9 simulation tasks from LIBERO and RoboCasa GR1, and Overall Avg. averages all 13 tasks.
Real-world task abbreviations: GMC: Grasp Middle Cup; GC: Grasp Coke; MT: Move Tide; SC: Stack Cups.
}
\tabcolsep 4pt
\label{tab:main}
\renewcommand{\arraystretch}{1.2}
\resizebox{\textwidth}{!}{%
\begin{tabular}{l cccc ccccc >{\columncolor{avgbg}}l cccc >{\columncolor{avgbg}}l}
\toprule
\multirow{2}{*}{Method}
  & \multicolumn{4}{c}{LIBERO}
  & \multicolumn{5}{c}{RoboCasa GR1}
  & \multicolumn{1}{c}{\multirow{2}{*}{\makecell{Sim\\Avg.}}}
  & \multicolumn{4}{c}{Real-world Franka}
  & \multicolumn{1}{c}{\multirow{2}{*}{\makecell{Overall\\Avg.}}} \\
\cmidrule(lr){2-5} \cmidrule(lr){6-10} \cmidrule(lr){12-15}
  & Spatial & Object & Goal & Long
  & B2C & C2B & Pm2B & Pl2B & T2C
  & \multicolumn{1}{c}{} & GMC & GC & MT & SC & \multicolumn{1}{c}{} \\
\midrule
GR00T~\cite{bjorck2025gr00t}
  & 91.2 & 97.6 & 92.2 & 88.6
  & 56 & 38 & 20 & 8 & 50
  & 60.2
  & 45 & 15 & 20 & 10 & 48.6 \\
$\pi_{0.5}$~\cite{pi05}
  & 95.6 & 98.4 & 95.8 & 91.0
  & 62 & 46 & 46 & 28 & 32
  & 66.1
  & 60 & 30 & 15 & 0 & 53.8 \\
\midrule
No Action Prior
  & 96.4 & 97.6 & 96.6 & 90.2
  & 62 & 46 & 28 & 18 & 44
  & 64.3
  & 65 & 5 & 45 & 25 & 55.3 \\
Action-State Prior
  & 97.0 & 97.2 & 97.6 & 91.0
  & 68 & 40 & 42 & 22 & 44
  & 66.5\tabgain{2.2}
  & 75 & 35 & 60 & 75 & 64.9\tabgain{9.6} \\
Action-State Prior + History
  & 97.6 & 99.0 & 96.6 & 96
  & 56 & 50 & 50 & 28 & 46
  & 68.8\tabgain{4.5}
  & 75 & 50 & 60 & 80 & 68.0\tabgain{12.7} \\
\bottomrule
\end{tabular}%
}
\vspace{-0.1in}
\end{table*}

\subsection{Qualitative Analysis of Long-tail Real-world Behavior}
\label{sec:qualitative}

\begin{figure*}[!t]
\centering
\includegraphics[width=0.99\textwidth]{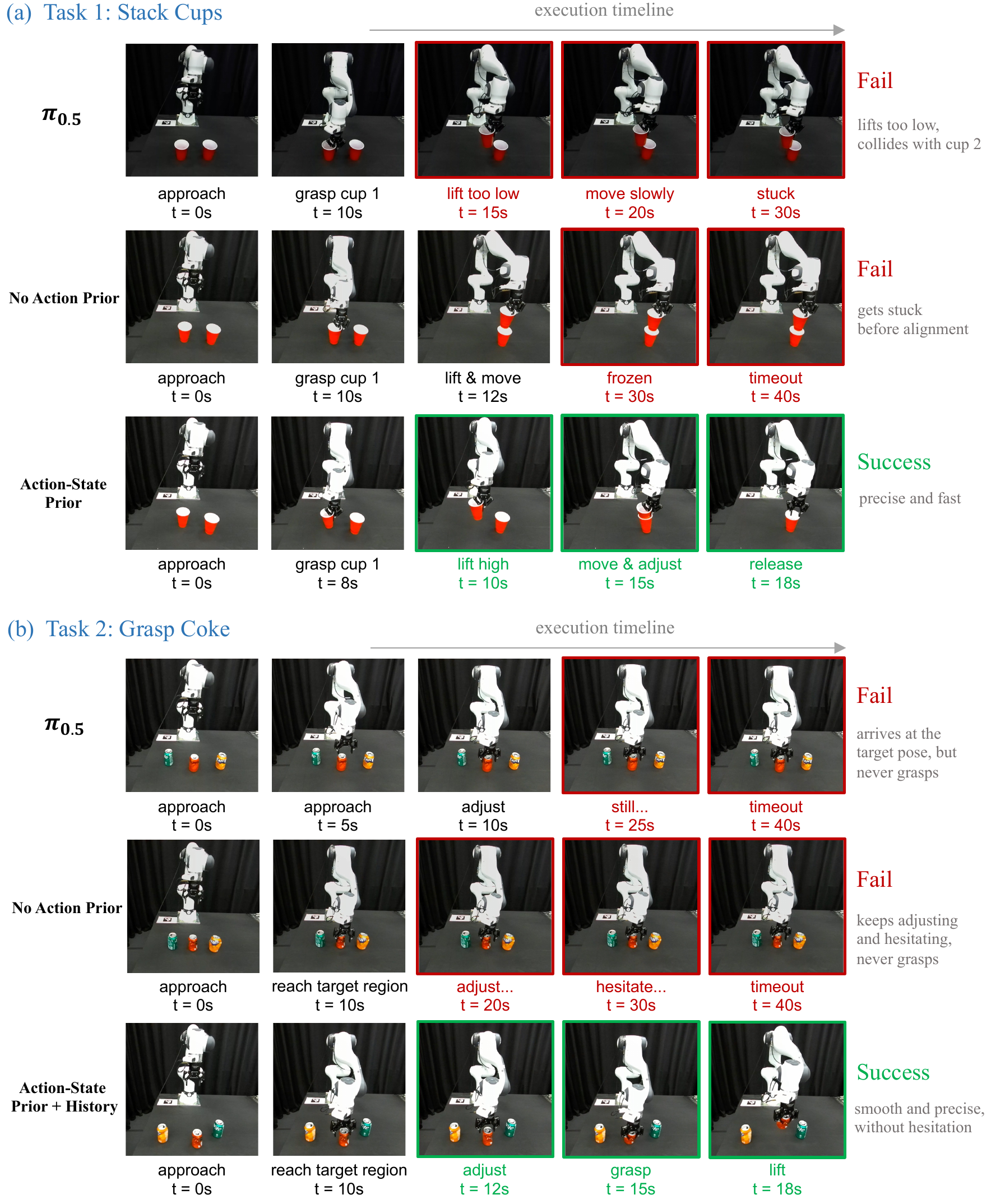}
\vspace{0.1in}
\caption{Qualitative comparison on two long-tail real-world tasks: (a) Stack Cups and (b) Grasp Coke. Rows show representative key frames for each method.
Without an action prior, policies move unstably.
$\pi_{0.5}$ underfits the real-world tail despite strong simulation results.
History compression reduces hesitation near the grasp pose.}
\label{fig:rollouts}
\end{figure*}

To understand why the action prior helps most on the data-scarce real-world tasks, we manually inspect the rollouts of all methods.
Fig.~\ref{fig:rollouts} visualizes representative key frames on the two most challenging long-tail tasks, Stack Cups and Grasp Coke.
The comparison reveals three characteristic behavior patterns that are not visible from success rates alone.

\vspace{0.05in}
\noindent\textbf{Without the action prior, the policy moves as if unfamiliar with its own arm.}
On the data-scarce real-world tasks, the No Action Prior policy frequently produces jerky and unstable motions.
In Fig.~\ref{fig:rollouts}(a), for example, it grasps the first cup within 10 seconds but then freezes before aligning it with the second cup, remaining stuck until timeout.
With the action prior, the motions become fast and precise, and the tasks are completed in a stable manner.
We attribute this difference to the division of labor in our two-stage design.
The action prior stage has already fitted the action distribution of the dataset efficiently.
The VLA stage therefore mainly learns to match the given task with motion patterns that the action module has mastered, leaving little hesitation during execution.

\vspace{0.05in}
\noindent\textbf{$\pi_{0.5}$ fits the simulated tasks well but underfits the long-tail real-world tasks.}
As shown in Fig.~\ref{fig:rollouts}(a), $\pi_{0.5}$ never lifts the first cup high enough, so it always collides with the second cup and fails on every Stack Cups trial.
A similar pattern appears on Grasp Coke, where it arrives at the target pose but never issues the grasp action (Fig.~\ref{fig:rollouts}(b)).
Considering its excellent performance on the simulation suites, we believe $\pi_{0.5}$ allocates its capacity to the data-rich simulated tasks while learning the long-tail real-world tasks insufficiently.

\vspace{0.05in}
\noindent\textbf{History compression enlarges the action receptive field and removes hesitation at decision points.}
Grasp Coke illustrates this effect most clearly.
In many failed rollouts of the history-free variants, the gripper aligns well with the bottle but never commits to the grasp.
This failure is closely related to the teleoperated demonstrations: before executing a grasp, the human operator often briefly pauses to align the gripper with the object.
A policy with a short temporal window can misinterpret this alignment phase as requiring further correction, causing it to stall near the grasp pose.
As shown in Fig.~\ref{fig:rollouts}(b), the No Action Prior policy reaches the target region at 10 seconds, yet it keeps making small adjustments around the aligned pose until timeout.
By compressing past states and actions into a single latent token, the learned encoder lets the VLM infer whether the robot is still approaching, aligning, or ready to grasp.
Equipped with this context, the Action-State Prior + History policy reaches the target region at the same time but commits to the grasp within 5 seconds, completing the task decisively.
This decisive behavior raises Grasp Coke from 35\% to 50\%, showing that the encoder effectively uses history.

\begin{figure*}[!t]
  \centering
  \includegraphics[width=0.99\textwidth]{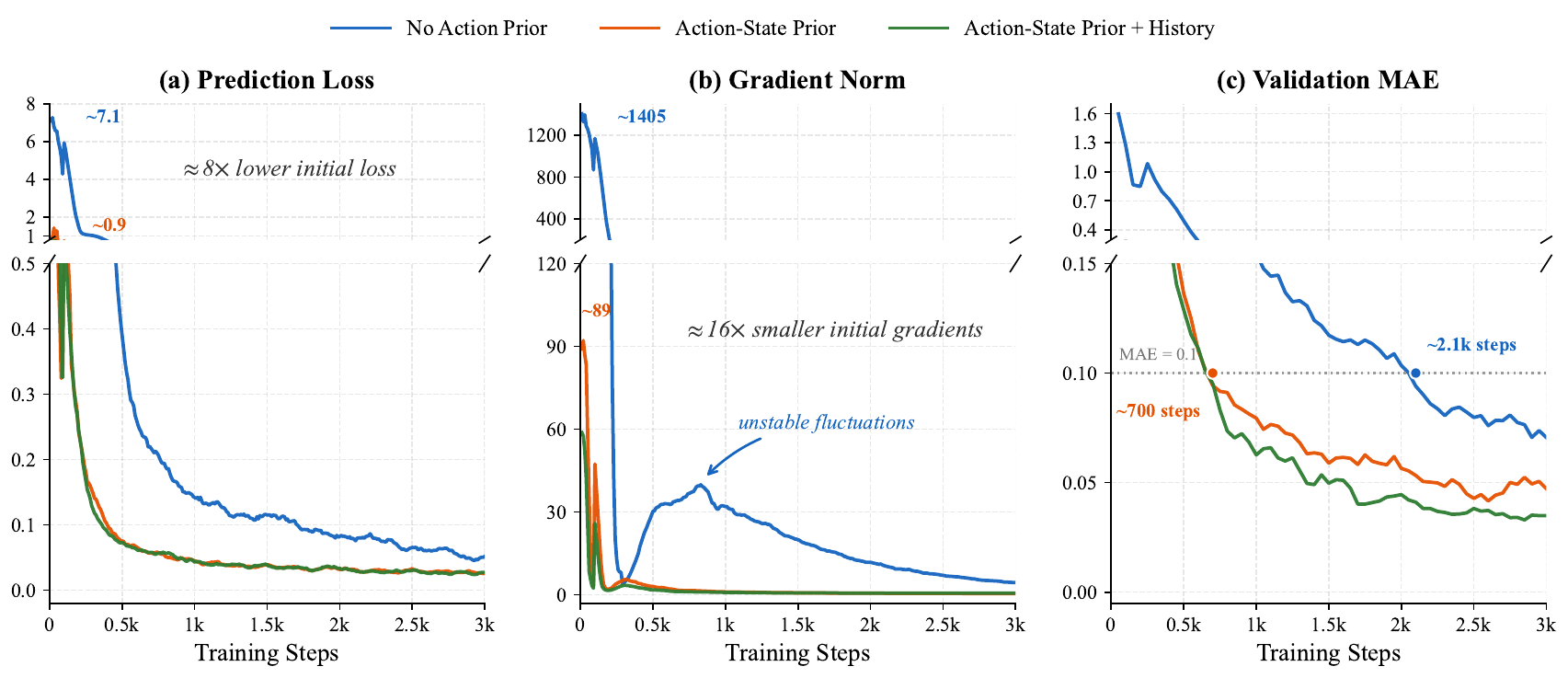}
  \caption{Training dynamics during the first 3{,}000 VLA training steps for three variants: No Action Prior, Action-State Prior, and Action-State Prior + History.
  From left to right: prediction loss, gradient norm, and MAE validation score.
  The y-axis of each plot is broken to simultaneously show the large initial gap and the fine-grained later behavior.
  With action prior learning, the initial prediction loss is reduced by nearly $8\times$ (from $\sim$7.1 to $\sim$0.9), and the initial gradient norm is reduced by nearly $16\times$ (from $\sim$1{,}400 to $\sim$90), leading to significantly faster and more stable convergence.}
  \label{fig:training_dynamics}
\vspace{-0.1in}
\end{figure*}

\subsection{Analysis of Training Dynamics and Optimization Stability}

Beyond final success rates and rollout behavior, we further examine how action prior learning changes the early optimization process of VLA training.
The key question is whether Stage~1 action prior learning can reduce the burden of Stage~2 VLA training, where the model must align visual-language features with continuous action generation.
Fig.~\ref{fig:training_dynamics} compares the first 3{,}000 Stage~2 steps of three variants under the same data, hyperparameters, and VLM backbone, isolating the effect of action prior initialization and history conditioning.

\vspace{0.05in}
\noindent\textbf{Stable initialization.}
The no-prior model starts from a poorly calibrated action module and suffers a much larger initial prediction loss.
In contrast, prior-based variants begin from a decoder that already models feasible motion, reducing the initial loss by nearly $8\times$ and reaching low loss much earlier.
This suggests that Stage~1 handles the low-level action calibration before cross-modal alignment begins.
Stage~2 can therefore focus less on discovering valid motion patterns from scratch and more on adapting these patterns to visual-language conditions.

\vspace{0.05in}
\noindent\textbf{Smoother gradients.}
Random action initialization also produces large and recurrent gradient spikes, while the learned decoder keeps gradients small and stable from the beginning.
This matters because stable action gradients reduce interference with the already useful representations in the VLM backbone.
% The optimization trajectory is consequently less dominated by action-space correction and more focused on task-conditioned alignment.

\vspace{0.05in}
\noindent\textbf{Faster validation convergence.}
The MAE curve confirms that the gain is not limited to training loss.
Prior-based variants reach the same validation error much earlier, indicating that the learned action prior also serves as an optimization prior for cross-modal policy learning.
The history variant further reaches the lowest MAE after the initial transient, consistent with its stronger downstream performance in Table~\ref{tab:main}.

\vspace{0.05in}
\noindent\textbf{Training efficiency.}
This acceleration is especially meaningful because Stage~1 is much cheaper than Stage~2 VLA training.
As summarized in Table~\ref{tab:hyperparams}, action prior learning uses a smaller backbone, processes no image or language tokens, supports an $8\times$ larger batch size, and runs for only 5k steps.
On the same 8$\times$H200 GPUs, this stage takes about 2 hours, compared with 20 hours for the 50k-step VLA training run.
This corresponds to only about 10\% additional wall-clock cost relative to the downstream VLA training stage.
Thus, Stage~1 is not another expensive VLA training phase; it is a lightweight action-only calibration stage that removes a large part of the early optimization burden before cross-modal learning begins.

\subsection{Ablation Study}
\label{sec:ablation}

\noindent We isolate three design choices behind the proposed framework:
\begin{enumerate}
    \item[\textbf{Q1.}] \textit{Does incorporating proprioceptive state into the action encoder improve the learned action prior?}
    \item[\textbf{Q2.}] \textit{Does history integration help, and does the learned action encoder compress history more effectively than naive MLP projection?}
    \item[\textbf{Q3.}] \textit{How long should the latent embedding distillation last?}
\end{enumerate}

\vspace{0.05in}
\noindent\textbf{Implementation Details.}
All ablations are conducted on the 9 simulation tasks using the same Stage~2 data, VLM backbone, and training configuration as the main comparison (Table~\ref{tab:hyperparams}).
Rows~2--5 in Table~\ref{tab:ablation_main} add Stage~1 action prior learning with different encoder inputs and history settings; history-enabled variants use a 15-step history horizon and $N_{decay}=5{,}000$.

\begin{table*}[!t]
\centering
\caption{Ablation of action/state prior learning and history integration on 9 simulation tasks.
Rows compare whether Stage~1 uses action-only or action-state trajectories, and whether Stage~2 receives encoder-compressed history context.
Success rates (\%) are reported for LIBERO and RoboCasa, with Avg. computed across all tasks.
Red numbers in the Avg. column denote absolute gains over the No Action Prior baseline.}
\label{tab:ablation_main}
\resizebox{\textwidth}{!}{%
\begin{tabular}{ccccccccccccc>{\columncolor{avgbg}}l}
\toprule
\multirow{2}{*}{\#}
  & \multicolumn{2}{c}{Prior Type} & {VLA Training}
  & \multicolumn{4}{c}{LIBERO} & \multicolumn{5}{c}{RoboCasa} & \multicolumn{1}{c}{} \\
\cmidrule{2-13}
% \cmidrule(lr){2-3}\cmidrule(lr){4-4}\cmidrule(lr){5-8}\cmidrule(lr){9-13}
  & Action & State & History
  & Spatial & Object & Goal & Long
  & B2C & C2B & Pm2B & Pl2B & T2C & \multicolumn{1}{c}{Avg.} \\
\midrule
1 &            &            &            & 96.4 & 98.2 & 97.2 & 87.2 & 60 & 38 & 30 & 30 & 44 & 64.6 \\
2 & \checkmark &            &            & 97.2 & 99.0 & 96.6 & 87.4 & 72 & 36 & 34 & 24 & 42 & 65.4\tabgain{0.8} \\
3 & \checkmark & \checkmark &            & 97.2 & 97.4 & 96.4 & 89.2 & 62 & 48 & 40 & 44 & 40 & 68.3\tabgain{3.7} \\
4 & \checkmark &            & \checkmark & 96.6 & 97.6 & 94.0 & 93.2 & 70 & 56 & 50 & 18 & 52 & 69.7\tabgain{5.1} \\
5 & \checkmark & \checkmark & \checkmark & 98.0 & 98.0 & 96.2 & 93.4 & 76 & 56 & 50 & 32 & 44 & 71.5\tabgain{6.9} \\
\bottomrule
\end{tabular}%
}
\end{table*}

\vspace{0.05in}
\noindent\textbf{Effect of Action Encoder Input.}
Adding proprioceptive state to the action encoder improves the average success rate from 65.4\% to 68.3\% (Rows~2--3 in Table~\ref{tab:ablation_main}).
The gain is most visible on RoboCasa, where the GR1 humanoid has a high-dimensional state space: Pl2B improves by 20 points, C2B by 12 points, and Pm2B by 6 points.
The same trend holds when history is enabled, with Row~5 outperforming Row~4 by 1.8\%.
This suggests that state information anchors relative action deltas to the robot's absolute body configuration.
Actions alone describe how the end effector or joints should change, but the same delta can correspond to different physical motions depending on the current pose.
By interleaving state and action tokens, the encoder observes both the local motion and the kinematic context from which it is executed.
This is especially useful for complex embodiments such as GR1, where body configuration provides important cues for disambiguating similar action chunks.

\vspace{0.05in}
\noindent\textbf{History context improves temporally extended manipulation.}
History integration further improves the action-state prior from 68.3\% to 71.5\% (Rows~3 and~5).
The benefit is strongest on sequential or multi-stage tasks, including LIBERO Long (+4.2 points) and several RoboCasa tasks such as B2C (+14), Pm2B (+10), and C2B (+8).
The action-only prior shows a similar trend, with Row~4 outperforming Row~2 by 4.3\%.
These improvements support the intuition that a single observation-action pair can be insufficient for temporally extended manipulation.
Historical state-action context exposes whether the policy has just approached, adjusted, or already committed to a subgoal, allowing the VLA model to make more stable decisions.

\begin{table}[t]
  \centering
  \caption{Comparison of history integration strategies.
  Ours uses the learned action encoder for history compression.}
  \label{tab:history_method}
  \tabcolsep 1pt
  \renewcommand{\arraystretch}{1.2}
  \resizebox{\columnwidth}{!}{%
  \begin{tabular}{lccccccccc>{\columncolor{avgbg}}l}
  \toprule
  \multirow{2}{*}{Method}
    & \multicolumn{4}{c}{LIBERO} & \multicolumn{5}{c}{RoboCasa} & \multicolumn{1}{c}{} \\
  \cmidrule(lr){2-5}\cmidrule(lr){6-10}
    & Spatial & Object & Goal & Long
    & B2C & C2B & Pm2B & Pl2B & T2C & \multicolumn{1}{c}{Avg.} \\
  \midrule
  No History & 97.2 & 97.4 & 96.4 & 89.2 & 62 & 48 & 40 & 44 & 40 & 68.3 \\
  Naive MLP     & 97.0 & 97.8 & 95.6 & 93.2 & 74 & 46 & 42 & 32 & 38 & 68.4\tabgain{0.1} \\
  Ours & \textbf{98.0} & \textbf{98.0} & \textbf{96.2} & \textbf{93.4} & \textbf{76} & \textbf{56} & \textbf{50} & \textbf{32} & \textbf{44} & \textbf{71.5}\tabgain{3.2} \\
  \bottomrule
  \end{tabular}%
  }
  \end{table}

\vspace{0.05in}
\noindent
\textbf{Encoder-based history compression is more effective than naive projection.}
Table~\ref{tab:history_method} verifies that the improvement comes from structured history compression rather than merely exposing the VLM to more numerical tokens.
Naive MLP projection barely improves over no history (68.4\% vs.\ 68.3\%), suggesting that raw historical state-action vectors are not easy for the VLM to interpret directly.
In contrast, the learned action encoder compresses history into the same latent space used for action prior learning and reaches 71.5\%.
The gap is most visible on RoboCasa, where the encoder improves over the naive baseline by 10 points on C2B, 8 points on Pm2B, and 6 points on T2C.
This indicates that the action encoder does more than reduce dimensionality: it converts raw temporal traces into structured motion embeddings that are easier to fuse with visual-language features.
Overall, the full model (Row~5) improves by 6.9\% over the no-prior baseline (Row~1), showing that state conditioning and encoder-based history compression are complementary.
From an architecture perspective, this insertion strategy also keeps the VLA interface compact.
Instead of appending long raw history tokens and asking the VLM to infer temporal structure from scratch, the encoder provides a single motion-aware summary token.
This makes history conditioning easy to reuse across embodiments, because the VLM receives a fixed-size temporal summary rather than embodiment-specific raw trajectories.
This is consistent with the role of the action prior: Stage~1 action-only learning first organizes motion history into a latent representation, and Stage~2 only needs to condition that representation on images and language.

\begin{figure}[!t]
  \centering
  \includegraphics[width=0.99\linewidth]{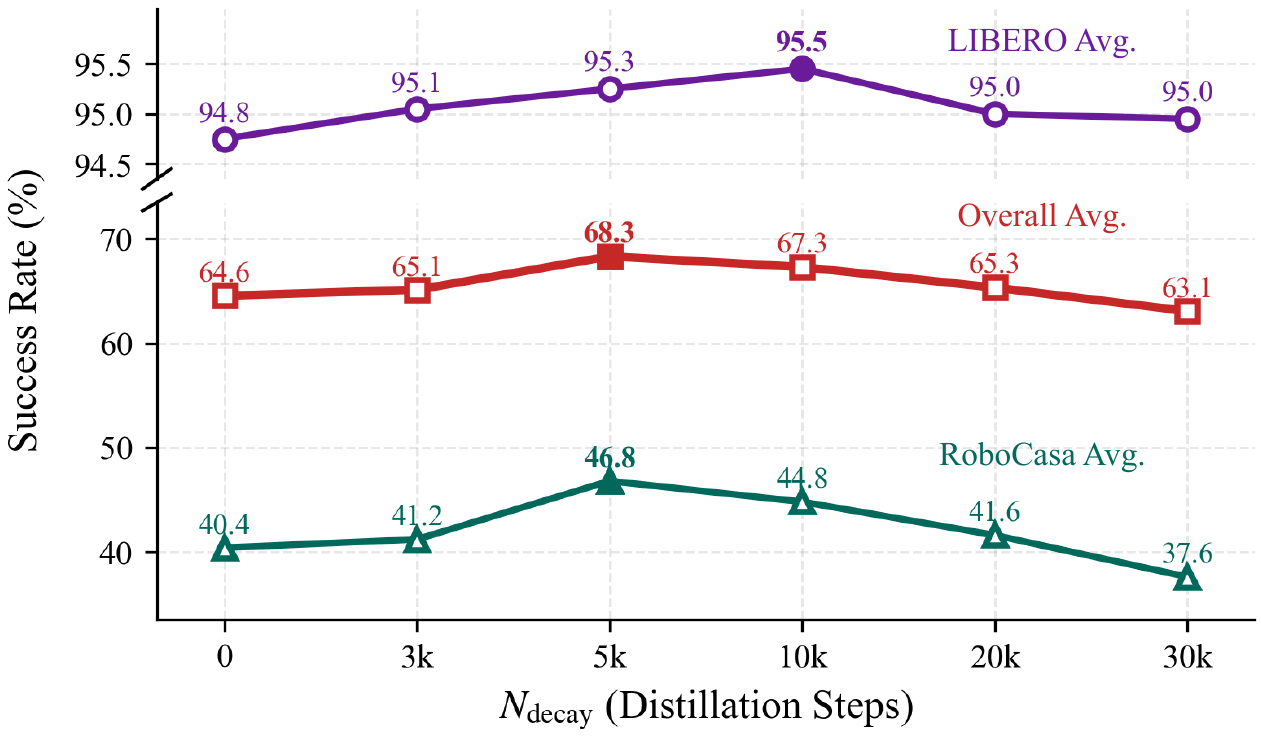}
  \caption{Effect of the distillation horizon under a fixed 30k-step VLA budget.
  Overall and RoboCasa performance peak at $N_{decay}=5{,}000$, while LIBERO peaks at 10k but remains nearly saturated.
  Too-short distillation underuses the prior, whereas too-long distillation constrains VLM refinement.}
  \label{fig:distillation}
\end{figure}

\vspace{0.05in}
\noindent\textbf{Effect of Distillation Steps.}
With a fixed 30k-step VLA budget, $N_{decay}$ controls how long the alignment and reconstruction losses guide the VLM.
Fig.~\ref{fig:distillation} shows a clear trade-off.
Without distillation ($N_{decay}=0$), the model corresponds to the no-prior baseline and reaches 64.6\%.
Activating distillation for 3k steps already improves the average to 65.1\%, and the best overall result appears at 5k steps with 68.3\%.
The gain is mainly driven by RoboCasa, where performance increases from 40.4\% to 46.8\%.
LIBERO is less sensitive because its average already stays around 95\%, although its best point appears at 10k steps.

This trend reflects the role of latent distillation in our training design.
During early Stage~2 training, $\mathcal{L}_{align}$ anchors the VLM-predicted action embedding to the structured latent space learned by the action encoder, providing a stable bridge from visual-language features to action generation.
However, the VLM eventually needs to refine representations using visual observations and language instructions that are absent from the action-only encoder.
When distillation is kept for too long, the VLM is forced to stay close to a narrower state-action representation, and full-training distillation drops to 63.1\%, below the no-prior baseline.
The best strategy is therefore to use the action prior as early guidance and then release the constraint for end-to-end refinement.
In this sense, $N_{decay}=5{,}000$ provides a practical balance: it gives the VLM enough latent guidance to enter the action space smoothly, while leaving most of Stage~2 for task-conditioned visual-language refinement.

% \begin{table*}[!t]
% \caption{Effect of distillation steps $N_{decay}$ on VLA performance.
% The total VLA training steps are fixed at 30{,}000. Best average is \textbf{bolded}.}
% \label{tab:distillation}
% \centering
% \renewcommand{\arraystretch}{1.1}
% \resizebox{0.75\textwidth}{!}{%
% \begin{tabular}{@{}c cccc ccccc c@{}}
% \toprule
% \multirow{2}{*}{$N_\text{decay}$}
%   & \multicolumn{4}{c}{LIBERO}
%   & \multicolumn{5}{c}{RoboCasa}
%   & \multirow{2}{*}{Avg.} \\
% \cmidrule(lr){2-5} \cmidrule(lr){6-10}
%   & Spatial & Object & Goal & Long
%   & B2C & C2B & Pm2B & Pl2B & T2C & \\
% \midrule
% 0 & 96.4 & 98.2 & 97.2 & 87.2 & 60 & 38 & 30 & 30 & 44 & 64.6 \\
% 3k  & 97.8 & 98.2 & 97.0 & 89.8 & 64 & 44 & 30 & 18 & 46 & 65.0 \\
% \textbf{5k}  & 97.2 & 97.4 & 97.2 & 89.2 & 62 & 48 & 40 & 44 & 40 & \textbf{68.3} \\
% 10k & 94.8 & 98.6 & 96.0 & 92.4 & 60 & 60 & 42 & 24 & 38 & 67.3 \\
% 20k & 95.2 & 98.6 & 96.8 & 88.8 & 74 & 36 & 28 & 22 & 48 & 65.3 \\
% 30k & 96.6 & 98.6 & 95.8 & 89.2 & 54 & 46 & 40 & 12 & 36 & 63.1 \\
% \bottomrule
% \end{tabular}%
% }
% \end{table*}

\subsection{Scaling Action Prior Data and Training Efficiency}

\begin{figure}[t]
\centering
\includegraphics[width=0.99\linewidth]{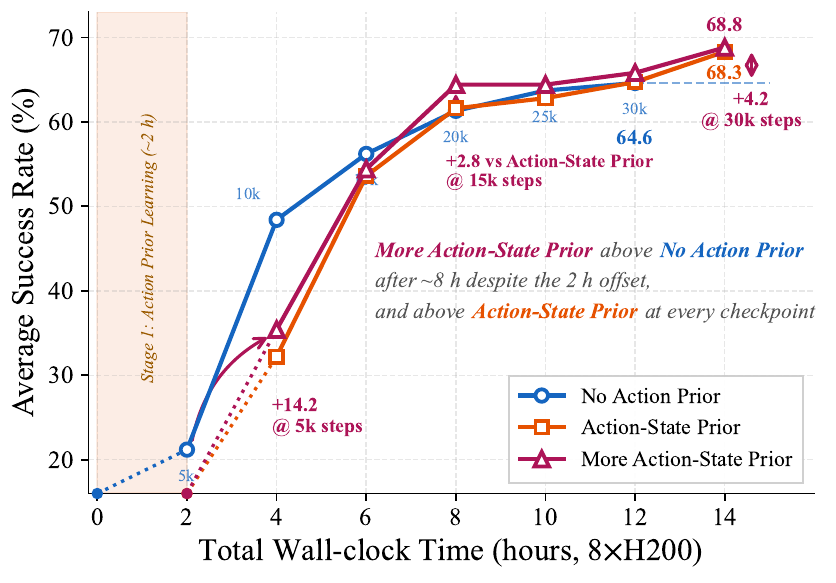}
\caption{Average success rate versus total wall-clock training time on 8$\times$H200 GPUs.
The prior variants include the 2-hour Stage~1 cost as an offset for fair accounting.
More Action-State Prior scales the Stage~1 action data while keeping the action-prior training configuration and cost unchanged.
It consistently outperforms the standard Action-State Prior throughout VLA training and achieves the best final success rate.
This shows that scaling action prior learning is meaningful: a broader action-only distribution transfers directly into faster and stronger downstream VLA training.}
\label{fig:convergence}
\end{figure}

\vspace{0.05in}
\noindent\textbf{Setup.}
Finally, we examine whether action prior learning benefits from more Stage~1 data and whether the extra Stage~1 cost is justified in wall-clock time.
This experiment changes only the data used to train the action prior.
The Stage~2 VLA data, VLM backbone, training steps, and all hyperparameters remain unchanged, so any downstream improvement must come from a stronger action prior rather than more VLA supervision.
Importantly, the Stage~1 configuration is also kept fixed: we use the same backbone, batch size, learning rate, and 5k training steps as in Table~\ref{tab:hyperparams}.
We only increase the amount of action data, rather than training the action module for more steps, because the action-only learning stage converges rapidly under large-batch optimization.
Specifically, we augment Stage~1 with LIBERO-90~\cite{liu2023libero} and the full 1{,}000 demonstrations for each selected RoboCasa task, increasing the Stage~1 action-data mixture from 565k to roughly 2.3M transitions ($4\times$).

\vspace{0.05in}
\noindent\textbf{The Stage~1 cost is quickly amortized.}
Fig.~\ref{fig:convergence} includes the full 2-hour Stage~1 cost on the wall-clock axis, so the prior variants start with an explicit time offset.
Even under this accounting, Action-State Prior catches up with the no-prior baseline by around 8 hours and finishes higher.
This confirms that the extra Stage~1 cost is quickly amortized during Stage~2 training.
The reason is that Stage~1 processes only compact state-action tokens, without expensive image or language encoders, making the learning stage much cheaper than VLA training.
The curve therefore reinforces the efficiency contrast discussed above: the action prior adds a small upfront cost, but shortens the much more expensive VLA optimization process.

\vspace{0.05in}
\noindent\textbf{Scaling action-prior data further strengthens downstream VLA training.}
At equal VLA step counts, the action prior consistently improves performance.
At 5k steps, Action-State Prior reaches 32.2\%, compared with 21.2\% for No Action Prior.
Scaling the Stage~1 data further raises this early result to 35.4\%, showing that a richer action distribution provides a stronger initialization.
The advantage persists later in training: the scaled prior reaches 64.4\% by 15k steps, nearly matching the no-prior model's final 30k-step result, and finishes at 68.8\%.
These results indicate that action prior learning is not only efficient, but also scalable: adding action-only data strengthens the motion prior and transfers directly to downstream VLA performance.
Although our Stage~1 corpus is still modest in scale, this trend suggests a clear path toward training action priors on broader action-only robot datasets, where more diverse motion distributions may yield stronger priors that could be transferred to the following VLA training.

\section{Conclusion}

We presented a two-stage training framework that introduces a dedicated action prior learning stage before VLA training.
By training a flow-matching-based encoder-decoder action module solely on unconditioned action trajectories, we equip the action head with structured knowledge of the action distribution before it encounters visual-language inputs.
An early-stage latent alignment distillation strategy further accelerates cross-modal convergence by anchoring the VLM to the learned action embedding space.
Experiments across 13 cross-embodiment tasks validate that the proposed approach yields faster convergence, higher success rates, and effective history compression at negligible cost.
We further demonstrate that enriching the Stage~1 data produces a stronger action prior that directly transfers to improved downstream VLA performance, confirming the scalability of the proposed framework.

\bibliographystyle{IEEEtran}
\bibliography{reference}

% \appendices
% \section{Additional Experimental Details}
% \label{app:exp_details}
%
% \begin{table}[!t]
% \centering
% \caption{Training data statistics for the 13 cross-embodiment tasks. Training frames denote the total number of state-action transitions.}
% \label{tab:training_data}
% \renewcommand{\arraystretch}{1.15}
% \resizebox{\columnwidth}{!}{%
% \begin{tabular}{@{}llr@{}}
% \toprule
% Benchmark & Task & Training Frames \\
% \midrule
% \multirow{4}{*}{LIBERO}
%  & Spatial Task Suite & 52{,}970 \\
%  & Object Task Suite & 66{,}984 \\
%  & Goal Task Suite & 52{,}042 \\
%  & Long Task Suite & 101{,}469 \\
% \midrule
% \multirow{5}{*}{RoboCasa}
%  & Bottle to Cabinet Close & 71{,}341 \\
%  & Cuttingboard Box to Basket & 48{,}292 \\
%  & Placemat to Basket & 48{,}066 \\
%  & Place to Bowl & 41{,}518 \\
%  & Tray to Cardboard Box & 39{,}739 \\
% \midrule
% \multirow{4}{*}{Real-world}
%  & Grasp Middle Cup & 11{,}312 \\
%  & Grasp Coke & 9{,}408 \\
%  & Move Tide to the Front of Book & 10{,}935 \\
%  & Stack Cups & 11{,}312 \\
% \midrule
% \textbf{Total} & \textbf{13 tasks} & \textbf{565{,}388} \\
% \bottomrule
% \end{tabular}%
% }
% \end{table}

% \section{Additional Discussion}
% \label{app:discussion}

% \begin{IEEEbiographynophoto}{Jane Doe}
% Biography text here without a photo.
% \end{IEEEbiographynophoto}

% \begin{IEEEbiography}[{\includegraphics[width=1in,height=1.25in,clip,keepaspectratio]{fig1.png}}]{IEEE Publications Technology Team}
% In this paragraph you can place your educational, professional background and research and other interests.\end{IEEEbiography}

\end{document}